%% file: arxiv_article.tex
\documentclass[10pt]{article}    

\usepackage{cite} 
\usepackage{mathtools,amsfonts} 
\usepackage{color}  
\usepackage{subfigure}  
\usepackage{url}  
\usepackage{ifthen}  
\usepackage{multicol}   
\usepackage[utf8]{inputenc} 
\def\includegraphics{}

\newboolean{publ}

\begin{document}


\title{Particle methods enable fast and simple approximation of Sobolev gradients in image segmentation}
 

\author{Ivo F.~Sbalzarini, Sophie Schneider, Janick Cardinale \\
\textit{MOSAIC Group, Center of Systems Biology Dresden (CSBD),} \\
\textit{Max Planck Institute of Molecular Cell Biology and Genetics,} \\
\textit{Pfotenhauerstr.~108, 01307 Dresden, Germany.} \\
\textit{E-Mail: ivos@mpi-cbg.de.}
}


\maketitle


\begin{abstract}
Bio-image analysis is challenging due to inhomogeneous intensity distributions and high levels of noise in the images. Bayesian inference provides a principled way for regularizing the problem using prior knowledge. A fundamental choice is how one measures ``distances'' between shapes in an image. It has been shown that the straightforward geometric $L^2$ distance is degenerate and leads to pathological situations. This is avoided when using Sobolev gradients, rendering the segmentation problem less ill-posed. The high computational cost and implementation overhead of Sobolev gradients, however, have hampered practical applications. 

We show how particle methods as applied to image segmentation allow for a simple and computationally efficient implementation of Sobolev gradients. We show that the evaluation of Sobolev gradients amounts to particle--particle interactions along the contour in an image. We extend an existing particle-based segmentation algorithm to using Sobolev gradients. Using synthetic and real-world images, we benchmark the results for both 2D and 3D images using piecewise smooth and piecewise constant region models. 
 
The present particle approximation of Sobolev gradients is 2.8 to 10 times faster than the previous reference implementation, but retains the known favorable properties of Sobolev gradients. This speedup is achieved by using local particle--particle interactions instead of solving a global Poisson equation at each iteration. The computational time per iteration is higher for Sobolev gradients than for $L^2$ gradients. Since Sobolev gradients precondition the optimization problem, however, a smaller number of overall iterations may be necessary for the algorithm to converge, which can in some cases amortize the higher per-iteration cost.
\end{abstract}

\ifthenelse{\boolean{publ}}{\begin{multicols}{2}}{}


\section*{Introduction}
Computational analysis of microscopy images has become a key step in many biological studies. It enables processing ever-larger sets of images at high throughput, improves reproducibility, and enables image-based modeling and simulation of biological systems~\cite{Eils:2003,Myers:2012,Sbalzarini:2013}. Additionally, computational image analysis methods can sometimes detect signals that the human eye cannot see~\cite{Danuser:2011}. Biological microscopy image data, however, come with their own set of challenges: They are usually diffraction limited and recorded at low signal-to-noise ratios (SNR), in order to minimize photo-toxic effects in the sample. In addition, the trend is to acquire three or even higher-dimensional images using microscopy techniques such as Selective Plane Illumination Microscopy (SPIM)~\cite{Huisken:2004}. The noise in these images is often not Gaussian. In fluorescence microscopy, only the fluorescently labeled structures are visible in the image, whereas other structures that are present in the sample are not imaged. This limits system observability. For these and other reasons, bio-image analysis tasks tend to be highly ill-posed. 

The strong ill-posedness of bio-image analysis problems requires additional regularization. This can be done in a biologically meaningful way by including prior knowledge about the imaged system and the imaging process into the analysis algorithms. This prior knowledge constrains the set of possible solutions to biologically and physically feasible ones, hence regularizing the analysis task. Bio-image analysis aims to extract quantitative information from higher-dimensional low-signal images, exploiting strong prior knowledge~\cite{Myers:2012}.

Bayesian inference provides a principled way for including prior knowledge into image-analysis algorithms. In Bayesian image analysis, one aims to maximize the {\em{posterior}} probability of the analysis result to be correct, given the observed image (see Methods section). Using Bayes' formula, the posterior is expressed as the product of the {\em{likelihood}} of observing the image and the {\em{prior}} probability of the result (see Methods section). In image segmentation, deformable models provide for a straightforward implementation of Bayesian, or ``model-based'', methods~\cite{Zhang:2004a,Montagnat:2001}. In this class of methods, the contours of the objects in the image are represented using models from computational geometry, such as splines~\cite{Kass:1988}, level sets~\cite{Sethian:1999b}, or triangulated surfaces~\cite{Dufour:2011}. These models then deform and move over the image so as to maximize the posterior. The evolution is driven by an optimization algorithm. 

In segmentation of fluorescence microscopy images, the prior knowledge that has been used to improve the quality and robustness of the results includes: (1) the point-spread function of the microscope to describe how the image has been acquired and to yield ``deconvolving segmentations''~\cite{Helmuth:2009,Helmuth:2009a,Jung:2009,Cardinale:2012,Paul:2013}; (2) the statistical distribution of the noise in the image in order to produce optimally denoised segmentations~\cite{Lecellier:2010,Paul:2013}; (3) topological priors about the connectedness of regions in the image~\cite{Cardinale:2012}; (4) physical priors about the mechanics of the imaged objects~\cite{Cardinale:2009,Helmuth:2009,Helmuth:2009a}; (5) the expected shape of the imaged objects~\cite{Leventon:2000,Raviv:2004}; (6) the expected color, texture, or motion of the imaged objects~\cite{Brox:2010}, etc.

A fundamental choice in any Bayesian method is how one measures distances between different shapes or segmentations. This is required by the optimization algorithm in order to perturb the deformable model and quantify the magnitude of this perturbation. Defining the gradient of the posterior hence relies on defining an inner product in the set of perturbations of a deformable model. The most common choice is the geometric $L^2$-type inner product. This has, however, been shown to lead to a pathological metric in which the ``distance'' between {\em{any}} two curves is zero~\cite{Michor:2003,Yezzi:2005,Sundaramoorthi:2007a,Sundaramoorthi:2007}. As a result, optimizing a deformable model via a $L^2$-type gradient flow is very sensitive to noise and requires length or curvature regularization of the contour~\cite{Sundaramoorthi:2007a}. Another undesirable feature of $L^2$-gradient flows is that they do not distinguish between global (rigid-body) {\em{motion}} of the contour and local {\em{deformation}}~\cite{Sundaramoorthi:2007}. The pathological nature of the Riemannian metric induced by the $L^2$ inner product on the space of smooth curves is effectively avoided when using Sobolev gradients~\cite{Neuberger:1997}. This considers the Bayesian posterior to be an element of a Sobolev space~\cite{Charpiat:2005,Sundaramoorthi:2005}. 

In a Sobolev space, the definition of neighborhood is naturally adapted to the segmentation problem~\cite{Charpiat:2005,Sundaramoorthi:2005,Sundaramoorthi:2007,Sundaramoorthi:2007a,Renka:2009}. This leads to results that are more robust against noise, smoother, and allow for more ``natural'' contour perturbations. It is well known and has previously been demonstrated that Sobolev gradients are useful in a variety of segmentation and tracking algorithms, since they allow one to use model formulations that would be ill-posed or numerically intractable when using $L^2$-based gradients~\cite{Sundaramoorthi:2008}.

Here, we show how a recently introduced particle-based deformable model~\cite{Cardinale:2012} can be extended to Sobolev gradients. The particle-method nature of the algorithm enables a novel, simple and computationally efficient approximation of Sobolev gradients. We exploit the fact that pairwise particle--particle interactions amount to a discrete convolution~\cite{Hockney:1988,Eldredge:2002,Koumoutsakos:2005,Schrader:2010}. Since Sobolev gradients can be computed by convolution of the $L^2$ gradient with a decaying kernel function (see Eq.~(13) in Ref.~\cite{Sundaramoorthi:2007}), they can be approximated by local interactions between the contour particles within a certain cutoff radius. This effectively avoids solving a global Poisson equation at each iteration, as was previously necessary~\cite{Renka:2009}. In the present particle-based algorithm, Sobolev gradients incur virtually no additional implementation overhead, are computationally efficient, and can easily be parallelized. We provide the mathematical formalism for approximating Sobolev gradients using discrete particle methods, and we show that this approximation preserves the known qualitative properties of Sobolev gradient flows as compared to $L^2$ gradient flows. 

\subsection*{Particle-based image segmentation}

Deformable models consist of a geometry representation and an evolution law~\cite{Montagnat:2001}. The evolution law acts on the degrees of freedom of the geometry representation~\cite{Zhang:2004a,Montagnat:2001}. In Bayesian methods, this is done such as to maximize the posterior or, equivalently (according to a Boltzmann distribution), minimize the energy functional obtained by taking the negative logarithm of the posterior (see Methods section). Deformable models can be continuous or discrete. In either case they can be implicit (also called geometric models, such as level sets~\cite{Sethian:1999b}) or explicit (also called parametric models, such as splines~\cite{Kass:1988}). The relationship between explicit and implicit models has previously been studied~\cite{Xu-Jr.:2000}. In continuous deformable models, the contour is represented by a mathematical object that is a continuous function of space. The contour is hence not limited to the pixel grid of a digital image, but may also evolve in sub-pixel steps~\cite{Helmuth:2009,Helmuth:2009a,Paul:2013}. Discrete representations directly store region labels at grid nodes. Grid nodes usually coincide with pixels. This allows straightforward representation of multiple regions and efficient querying of spatial information. 

\begin{figure}
  \centering
    \def\svgwidth{0.3\columnwidth}      
    \footnotesize   
    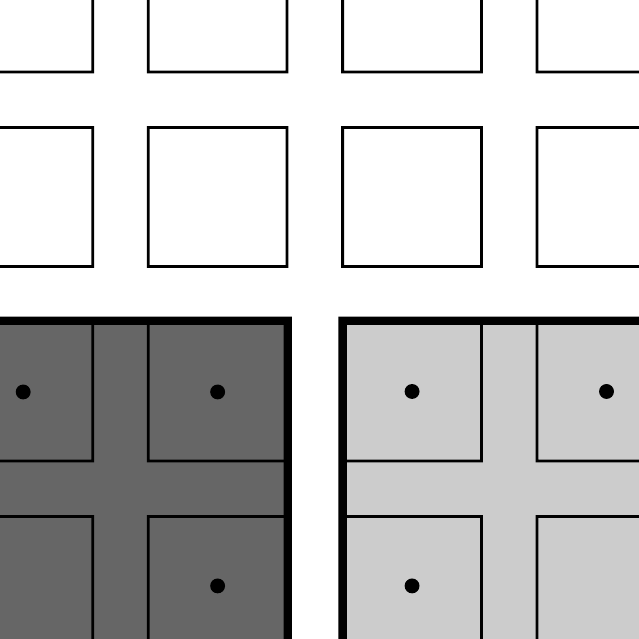
    \caption{Discrete contour representation using particles. Shown is the pixel grid with two closed foreground regions (dark and light gray) and the open background region (white). Dots represent particles marking discrete contour points. The contour pixels belong to the respective foreground region and contribute to its region statistics. The bold lines illustrate the  contour $\Gamma$ that is represented by the particles.}\label{fig:fg_simple}
\end{figure}

Here, we consider a discrete deformable model where the contour is represented by computational particles placed in the pixels around which the contour passes. This is illustrated in Fig.~\ref{fig:fg_simple} and provides a geometry representation that is somewhere between explicit and implicit~\cite{Shi:2008}. Particles migrate to neighboring pixels in order to deform the contour. All foreground regions are defined as closed sets, i.e., the contour pixels belong to the foreground region. There is only one single background region in the entire image, which is an open set. Foreground regions are constrained to be connected sets of pixels, whereas the background region may be disconnected. See Ref.~\cite{Cardinale:2012} for details. 

\subsection*{The Region Competition (RC) algorithm}

The Region Competition algorithm~\cite{Cardinale:2012} is a discrete optimization algorithm to drive the evolution of contours represented by discrete particles. The algorithm can advance an arbitrary number of contours simultaneously and provides topological control over the evolving contours. Topological control ensures that contours remain connected (i.e., ``intact'' according to the region definition used~\cite{Cardinale:2012}) and provides control over merging and splitting events between different contours. 
Local energy minimization is done by gradient descent, approximating the gradient as the energy difference between a perturbed and the original set of particles. The result of the algorithm does not depend on the order in which the particles are processed (or, equivalently, the indexing order of the particles), since the particles are ranked in order of decreasing energy difference before the moves are executed. 

The RC algorithm readily generalizes to 3D images, as the particle contour representation remains unaltered. RC has been demonstrated on 2D and 3D images using a variety of image models, including piecewise constant, piecewise smooth, and deconvolving models~\cite{Cardinale:2012}. The computational performance of RC is competitive with other state-of-the-art discrete methods, such as graph-cuts~\cite{Delong:2010}. For details, we refer to the original publication~\cite{Cardinale:2012}. 

The original work presenting the RC algorithm used a $L^2$-type discrete gradient approximation, which is known to frequently get trapped in local minima and be sensitive to noise. Here, we show how Sobolev gradients can be approximated in the same algorithm, and we demonstrate the computational efficiency of the resulting implementation. 

\subsection*{Sobolev active contours}
Sobolev active contours are deformable models whose evolution is driven by the Sobolev gradient flow of the underlying Bayesian energy~\cite{Charpiat:2005,Sundaramoorthi:2005,Sundaramoorthi:2007,Sundaramoorthi:2008,Renka:2009}. The original application of Sobolev gradients is the numerical solution of nonlinear partial differential equations~\cite{Neuberger:1997}. In image segmentation, Sobolev gradients have successfully been used for continuous deformable models, in particular for level sets methods. There, the Sobolev gradients are either directly computed on the continuous implicit function representation using variational calculus~\cite{Renka:2009}, or the implicit representation is intermediately translated to an explicit (linear spline) representation and the Sobolev gradients approximated there~\cite{Sundaramoorthi:2007}.

Here, we integrate Sobolev gradients into discrete contour models represented by particles. The RC algorithm amounts to a gradient descent, where the gradient is approximated at discrete points. The original work used a $L^2$-type gradient~\cite{Cardinale:2012}. This implies that one can only consider energies (image models) belonging to the $L^2$ inner-product function space. This inner product has a number of undesirable properties for deformable models~\cite{Sundaramoorthi:2007a}. For discrete models, the following two of these properties are of special interest:  First, the inner product does not contain any regularity terms. There is hence nothing in the metric that would discourage the emergence and evolution of non-smooth contour/time (hyper) surfaces. Hence, in the presence of noise, the contour becomes non-smooth during evolution, which in term reduces the numerical accuracy of the gradient approximation. Curvature regularization via priors is typically required to prevent this. Second, the $L^2$-type gradient is ignorant with respect to the type of contour motion, such as global translations or local deformations. Intuitively, the contour therefore locally optimizes ``on a small scale'' and frequently gets trapped in local optima of the Bayesian posterior.

A Sobolev function space is equipped with an inner product that contains $L^p$-terms and derivatives of the function. The metric on that space induced by this inner product hence includes smoothness terms that allow addressing the regularity issues mentioned above. Using such a metric does not affect the global minimum of the function, but it amounts to {\em{preconditioning}} the gradient flow, hence rendering it less ill-posed.

While Sobolev gradients are computationally more expensive to compute, they often result in a smaller overall number of iterations required by the segmentation algorithm to converge to a local optimum. Depending on the problem at hand, the cost of Sobolev gradient approximation may hence be amortized.  

\subsubsection*{Inner product}
We consider a Sobolev space $W^{1,2}$, which is a Hilbert space $H^1$\index{Hilbert space}, with inner product~\cite{Sundaramoorthi:2007}
\begin{equation}
  \label{eq:sobolev_inner_product}
\langle h,k \rangle_{H^1} := \bar{h}\cdot\bar{k}+\epsilon \cdot E^2\cdot\langle \nabla h,\nabla k\rangle_{L^2},
\end{equation}
where $h$ and $k$ are elements of the tangent space of the evolving contour $\Gamma$. The tangent space is the set of all possible deformations of $\Gamma$. The $\nabla$-operator in Eq.~(\ref{eq:sobolev_inner_product}) is with respect to the $L^2$-norm. The scalar $\epsilon\in\mathbb{R}^{+}$ is a hyper parameter for smoothness and $E\in\mathbb{R}^{+}$ determines the length scale of the smoothness terms in the inner product~\cite{Sundaramoorthi:2007}. The average $\bar{h}$ of $h$ over $\Gamma$ is
\begin{equation}
  \label{eq:avg_deformation}
\bar{h}=\frac{1}{|\Gamma|}\int_{\Gamma}  h(s)\, \mathrm{d}s \, , 
\end{equation}
where $s$ is the intrinsic position along the contour, in image coordinates (e.g., pixels). The average $\bar{k}$ is defined similarly. The $L^2$ inner product is
\begin{equation}
  \label{eq:L2_inner_prod}
  \langle h,k\rangle_{L^2}=\frac{1}{|\Gamma|}\int_{\Gamma}  h(s) k(s)\, \mathrm{d}s \, . 
\end{equation}

\section*{Results and Discussion}\label{sec:results}

We present a deformable model where the geometry representation uses a discrete particle method and the the evolution law is given by a Sobolev gradient flow. We use the RC algorithm to numerically optimize the resulting system, and we demonstrate and benchmark its behavior on synthetic and real-world images.

\subsection*{Particle approximation of Sobolev gradients}\label{sec:sobolev_gradients}
We adapt the RC algorithm to use an approximate Sobolev gradient flow to maximize the segmentation model posterior. In Bayesian energy minimization, as described in the Methods section, this requires computing the first-order Sobolev gradient $\nabla_{H^1}\mathcal{E}$ of the image energy $\mathcal{E}$ using the metric induced by the inner product in Eq.~(\ref{eq:sobolev_inner_product}). It has been shown~\cite{Sundaramoorthi:2007} that this amounts to a convolution of the $L^2$-type gradient $\nabla\mathcal{E}$ 
\begin{equation}
  \label{eq:sobolev_graident}
  \nabla_{H^1} \mathcal{E}(s) = \int_{\Gamma} \tilde{K}(\hat{s}-s)\cdot\nabla\mathcal{E}(\hat{s})\mathrm{d}\hat{s} = (\tilde{K}*\nabla\mathcal{E})(s)
\end{equation}
with convolution kernel
\begin{equation}
  \label{eq:sobolev_kernel}
  \tilde{K}(r) = \frac{1}{E}\left(1 + \frac{(|r|/E)^2-(|r|/E)+1/6}{2\epsilon}\right),\quad r\in\left[-\frac{E}{2},+\frac{E}{2}\right] \, .
\end{equation}
Figure~\ref{fig:sobolev_kernel} shows $\tilde{K}$ for different $\epsilon$~\cite{Sundaramoorthi:2007}. 
Equation (\ref{eq:sobolev_graident}) enables computing the first-order Sobolev gradient from the $L^2$ gradient. The convolution domain is $\Gamma$. All distances $(\hat{s}-s)$ are hence geodesic distances along the contour.  

  \subsection*{Figure 2 - Sobolev gradient kernel}
     Kernel $\tilde{K}$ for different $\epsilon$. The solid, dotted, and dashed curves show $\tilde{K}$ for $\epsilon=1/24$, $\epsilon=0.06$, and $\epsilon=0.08$, respectively. The kernel becomes local (decays to zero at $r=\pm E/2$) for $\epsilon=1/24$ .

\begin{figure}
  \centering
  \def\svgwidth{0.45\columnwidth}        
  \footnotesize       
  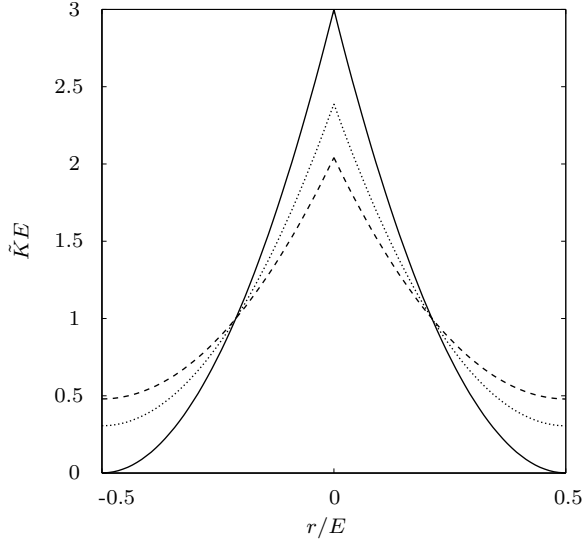 
  \caption{Sobolev gradient kernel $\tilde{K}$ for different $\epsilon$. The solid, dotted, and dashed curves show $\tilde{K}$ for $\epsilon=1/24$, $\epsilon=0.06$, and $\epsilon=0.08$, respectively. The kernel becomes local (decays to zero at $r=\pm E/2$) for $\epsilon=1/24$.}\label{fig:sobolev_kernel}
\end{figure}

\subsubsection*{Length scale}
\label{sec:sobolev_kernel_discussion}
The parameter $E$ controls the length scale of the inner product. It is often set equal to the total contour arc-length $|\Gamma|$ in order to obtain a scale-invariant inner product~\cite{Sundaramoorthi:2005,Sundaramoorthi:2007}. Gradient information is then shared along the entire contour, which enables global (rigid-body) contour movements, such as translations and rotations. However, this also introduces a global coupling into the computations, requiring the solution of a global Poisson equation~\cite{Renka:2009} or convolution~\cite{Sundaramoorthi:2005,Sundaramoorthi:2007} at each iteration of the algorithm, which is computationally expensive.

Here, we fix $E$ and thus impose a scale with respect to the image coordinate system. This makes sense in bio-image segmentation, where the length-scale of the objects of interest is often known. $E$ hence becomes a user-defined free parameter of the method. We find $E\approx 10\ldots 12$ to work well on the examples presented hereafter. We always use $\epsilon=\frac{1}{24}$ in order to get a kernel that decays to zero at $\pm E/2$, rendering  particle--particle interactions local (see Fig.~\ref{fig:sobolev_kernel}). Using this kernel, a length scale of $E$ of 10 to 12 hence corresponds to a particle--particle interactions radius of 5 to 6 pixel. As we show below, this leads to a simple and fast approximation of Sobolev gradients, which is accurate enough to retain the favorable properties of Sobolev gradient flows.

\subsubsection*{Discrete particle approximation}
In a discrete particle representation of the contour, the domain of the convolution in Eq.~(\ref{eq:sobolev_graident}) is the set of all particles representing that contour. In RC, the particles store a discrete $L^2$-type gradient approximation on both sides of the contour $\Gamma$, we can thus approximate the Sobolev gradient by discretizing Eq.~(\ref{eq:sobolev_graident}) over particles~\cite{Hockney:1988}. 

Let $\mathcal{Q}_p$ be the set of particles that are located in the support of $\tilde{K}$ and belong to the same contour as particle $p$, i.e., have region label $l=l_p$, thus: $\mathcal{Q}_p =\{q|\,d(x_q,x_p)<E/2, \, l_p = l_q\}$. Similarly, let $\mathcal{Q}'_p$ be the particles within the kernel support that lie on the other side of the contour, i.e., $\mathcal{Q}'_p =\{q|\,d(x_q,x_p)<E/2, \, l_p = l'_q\}$. For each particle $p$ we then compute the energy difference as:
\begin{equation}     
  \label{eq:smoothing_sobolev_energy_update} 
  \Delta\mathcal{E}_p \leftarrow  \frac{1}{|\mathcal{Q}_p|}\sum_{q\in\mathcal{Q}_p} \tilde{K}(d(x_q,x_p))\Delta\mathcal{E}_q - \frac{1}{|\mathcal{Q}'_p|}\sum_{q\in\mathcal{Q}'_p} \tilde{K}(d(x_q,x_p))\Delta\mathcal{E}_q \, . 
\end{equation}
This amounts to local pairwise particle--particle interactions, as they are commonplace in particle methods, where a convolution is approximated by a sum over particles~\cite{Hockney:1988,Eldredge:2002,Koumoutsakos:2005,Schrader:2010}.
We use cell lists~\cite{Hockney:1988} with a cell edge-length equal to the interaction cutoff radius of $E/2$ in order to efficiently find the neighbors (interaction partners) of each particle. This reduces the average time complexity of the discrete convolution from $O(N^2)$ to $O(N)$ for a total of $N$ contour particles. 

If both terms in Eq.~(\ref{eq:smoothing_sobolev_energy_update}) have the same sign, the approximated Sobolev gradients on both sides of the contour  have opposite directions. This happens at an extremum of the energy. Since the discrete contour representation does not allow sub-pixel deformations, the contour then oscillates. This, however, is easily detected~\cite{Cardinale:2012} and the optimizer switches back to using $L^2$-type gradients when oscillations occur.

According to Eq.~(\ref{eq:sobolev_graident}), the distance between two particles on the contour $\Gamma$ should be the geodesic arc distance $(\hat{s} - s)$ between the two points where the particles are located. The present discrete representation, however, does not allow computing this quantity. Since we consider a relatively small support of $\tilde{K}$ ($E/2\approx 5$ pixel), and objects in biological microscopy images are usually smoother than that, we approximate geodesic distances by Euclidean distances $d(\cdot,\cdot)$. This approximation breaks down when contours in the image significantly curve on length scales below $E$. One may then argue, however, that a continuous contour representation is anyway advised~\cite{Helmuth:2009,Helmuth:2009a}.

\subsection*{Benchmarks and comparisons}\label{sec:results}
We demonstrate the computational efficiency of particle-approximated Sobolev gradients in the RC algorithm~\cite{Cardinale:2012} by comparing with a previous mesh-based level-set implementation~\cite{Renka:2009}. We also illustrate that the present approximation retains the known favorable properties of Sobolev gradient flows by qualitatively comparing with segmentations obtained using approximated $L^2$ gradients for a range of regularization parameters $\lambda$ (see Methods section for an explanation of the meaning of $\lambda$). All benchmarks are performed using a C++ implementation of RC with $L^2$ and Sobolev gradients, run on a single 2.67\,GHz Intel i7 core with 4\,GB RAM using the Intel C++ compiler (v.~12.0.2).

\begin{figure}
\centering
\begin{minipage}{0.24\textwidth} 
  \includegraphics[width=\columnwidth]{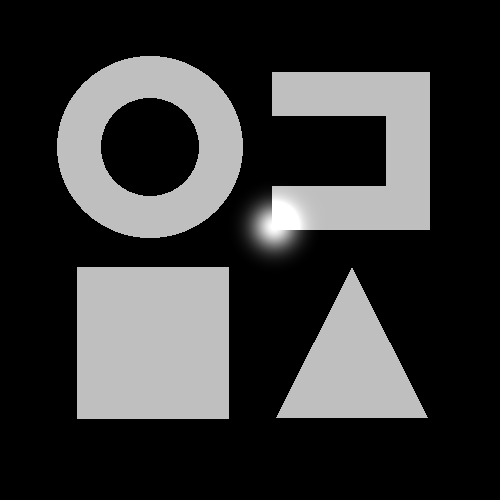} \centering A: ground truth
  \label{fig:fanta4_gt}
\end{minipage}
\begin{minipage}{0.24\textwidth} 
  \includegraphics[width=\columnwidth]{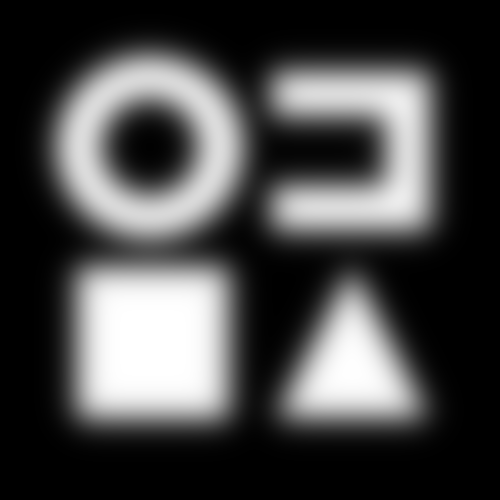} \centering B: blurred image
  \label{fig:fanta4_conv}
\end{minipage}
\begin{minipage}{0.24\textwidth} 
  \includegraphics[width=\columnwidth]{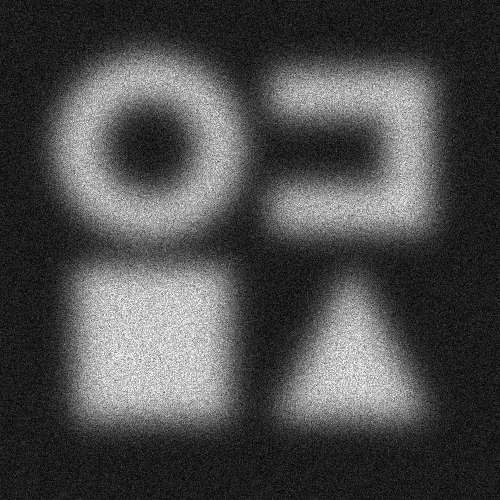} \centering C: noisy, blurred image
  \label{fig:fanta4_data}
\end{minipage}
\begin{minipage}{0.24\textwidth} 
  \includegraphics[width=\columnwidth]{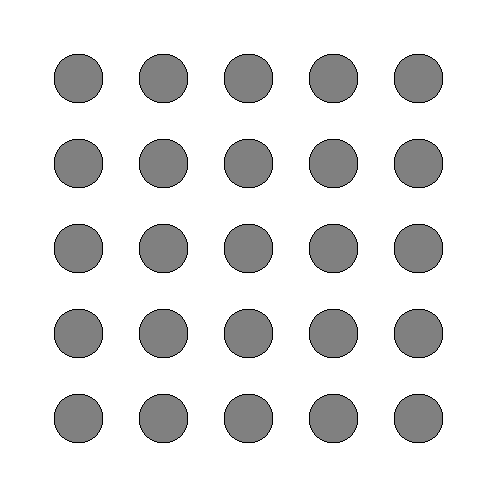} \centering D: initial segmentation
  \label{fig:fanta4_init}
\end{minipage}

\begin{minipage}{0.24\textwidth} 
  \includegraphics[width=\columnwidth]{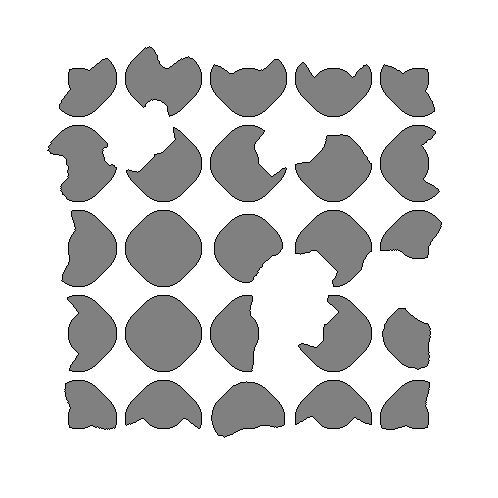} \centering E: $L^2$ after 15 iter.
  \label{fig:sobolev_without_frame15}
\end{minipage}
\begin{minipage}{0.24\textwidth} 
  \includegraphics[width=\columnwidth]{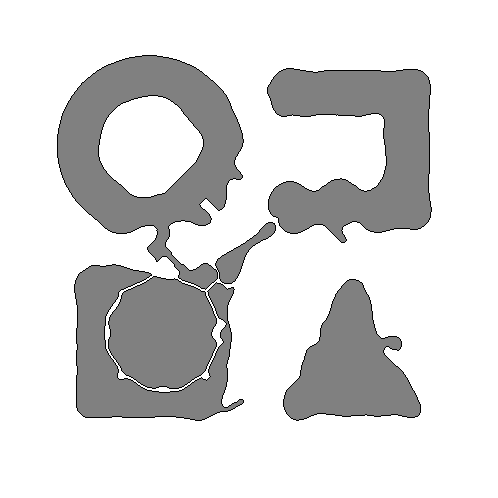} \centering F: $L^2$ after 59 iter.
  \label{fig:sobolev_without_frame59}
\end{minipage}
\begin{minipage}{0.24\textwidth} 
  \includegraphics[width=\columnwidth]{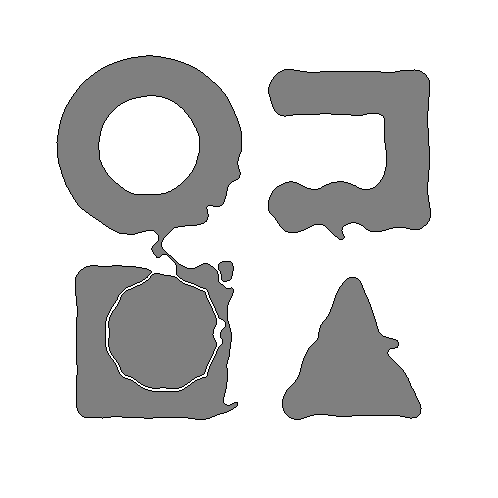} \centering G: $L^2$ after 88 iter.
  \label{fig:sobolev_without_frame88}
\end{minipage}
\begin{minipage}{0.24\textwidth} 
  \includegraphics[width=\columnwidth]{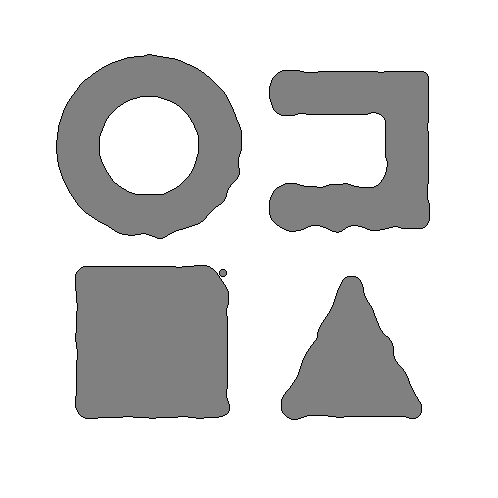} \centering H: $L^2$ after 500 iter.
  \label{fig:sobolev_without_frame500}
\end{minipage}

\begin{minipage}{0.24\textwidth} 
  \includegraphics[width=\columnwidth]{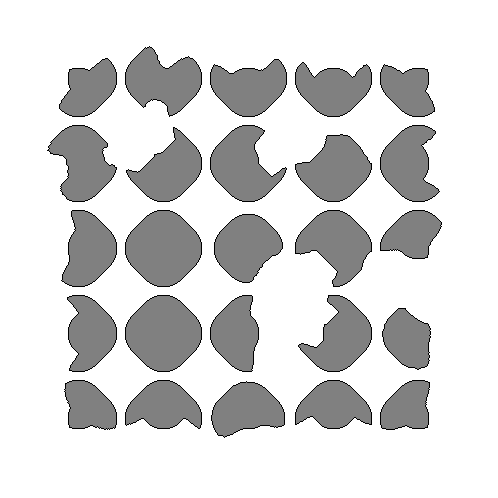} \centering I: Sobolev after 15 iter.
  \label{fig:sobolev_with_frame15}
\end{minipage}
\begin{minipage}{0.24\textwidth} 
  \includegraphics[width=\columnwidth]{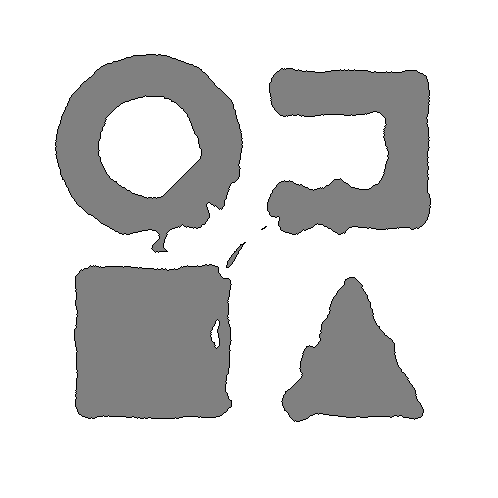} \centering J: Sobolev after 59 iter.
  \label{fig:sobolev_with_frame59}
\end{minipage}
\begin{minipage}{0.24\textwidth} 
  \includegraphics[width=\columnwidth]{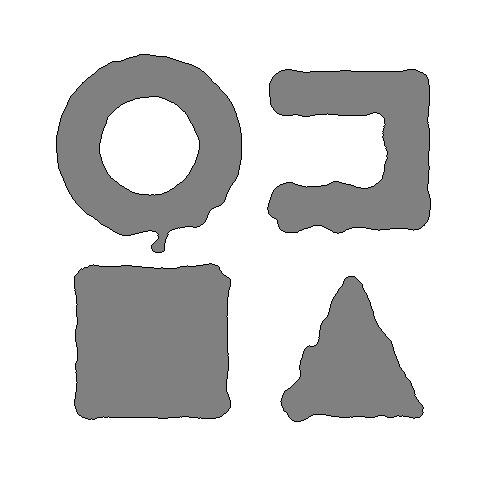} \centering K: Sobolev after 88 iter.
  \label{fig:sobolev_with_frame88}
\end{minipage}
\begin{minipage}{0.24\textwidth} 
  \includegraphics[width=\columnwidth]{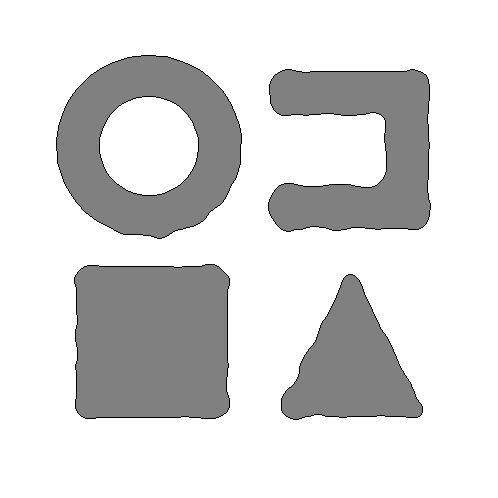} \centering L: Sobolev after 282 iter.
  \label{fig:sobolev_with_frame282}
\end{minipage}
\caption{Sobolev flows favor smooth contour evolution. We show the region evolution using $L^2$ and Sobolev gradient approximations for the piecewise smooth deconvolving energy described in Ref.~\cite{Cardinale:2012}. (A)~Ground-truth data with overlaid Gaussian point-spread function of width $\sigma=15$ pixels. (B)~Blurred image convolved with the point-spread function. (C)~Convolved image with Poisson noise of PSNR 6. (D)~Initialization for both algorithms. (E--H) RC segmentations after 15, 59, 88, and 500 iterations using $L^2$-type gradients. (I--L) RC segmentations after 15, 59, 88, and 282 iterations using Sobolev gradients. The Sobolev gradient flow is more regular and converges faster. The same $\lambda=0.04$ is used for both flows. The energy landscapes are hence identical, permitting direct comparison between optimization trajectories.}\label{fig:Sobolev_fantastic4}
\end{figure}

The metric induced by the inner product in a Sobolev space includes smoothness terms that favor smooth contour evolution, whereas the $L^2$-type gradient flow tends to produce non-smooth contours~\cite{Sundaramoorthi:2007,Sundaramoorthi:2008}. We use an artificial deconvolution problem to illustrate that the present particle approximation retains this property. Figure~\ref{fig:Sobolev_fantastic4}C shows the artificial data, which is generated by blurring the ground-truth scene in Fig.~\ref{fig:Sobolev_fantastic4}A with the point-spread function shown as a bright spot. This yields the noise-free image shown in Fig.~\ref{fig:Sobolev_fantastic4}B. The input image for RC is obtained by adding modulatory Poisson noise (Fig.~\ref{fig:Sobolev_fantastic4}C), modeling a common situation in fluorescence microscopy. From this image, the algorithm should reconstruct the denoised, deblurred, and segmented objects of the ground truth. 
The RC algorithm is always initialized with 25 bubbles on a Cartesian grid as show in Fig.~\ref{fig:Sobolev_fantastic4}D.
Intermediate results during energy minimization are shown in Figs.~\ref{fig:Sobolev_fantastic4}E--G and \ref{fig:Sobolev_fantastic4}I--K for $L^2$ and Sobolev gradient flows, respectively. During the first 15 iterations, regions evolve almost identically for both gradient types (Figs.~\ref{fig:Sobolev_fantastic4}E and I). After 59 iterations, the Sobolev-gradient approach is closer to the final solution (Figs.~\ref{fig:Sobolev_fantastic4}F and J). After 88 iterations, it starts oscillating and therefore falls back to the $L^2$-gradient mode (Figs.~\ref{fig:Sobolev_fantastic4}G and K). It converges after 282 iterations (Fig.~\ref{fig:Sobolev_fantastic4}L). The $L^2$-gradient approach converges after 500 iterations (Fig.~\ref{fig:Sobolev_fantastic4}H). The results for both algorithms are shown in Figs.~\ref{fig:Sobolev_fantastic4}H and \ref{fig:Sobolev_fantastic4}L. As expected, they are visually indistinguishable, because changing the gradient flow (at constant $\lambda=0.04$ for both flows) does not change the location of the energy minimum. However, the Sobolev flow uses a different optimization trajectory, which produces intermediate contours that are smoother than those generated by the $L^2$ flow. This leads to faster convergence and lower computational cost, since the the computational cost of the RC algorithm is proportional to the total contour length~\cite{Cardinale:2012}. The average iteration times are 0.74\,s with and 0.73\,s without Sobolev gradients. 

\begin{figure}
\begin{minipage}{0.24\textwidth} 
     \includegraphics[width=\columnwidth]{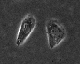}
     \includegraphics[width=\columnwidth]{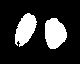} \centering A: TwoCells
\end{minipage}
\begin{minipage}{0.24\textwidth} 
    \includegraphics[width=\columnwidth]{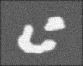} 
  \includegraphics[width=\columnwidth]{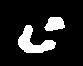}  \centering B: TwoObjects
\end{minipage}
\begin{minipage}{0.24\textwidth} 
    \includegraphics[width=\columnwidth]{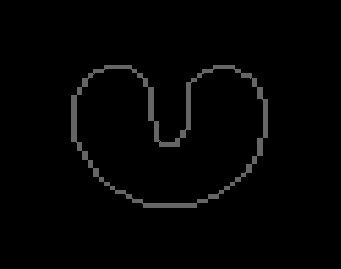}
    \includegraphics[width=\columnwidth]{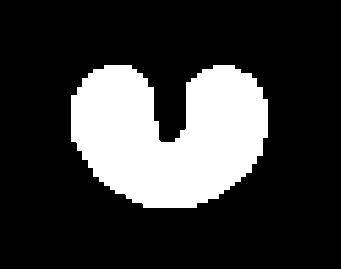}  \centering C: Horseshoe
\end{minipage}
\begin{minipage}{0.24\textwidth} 
    \includegraphics[width=\columnwidth]{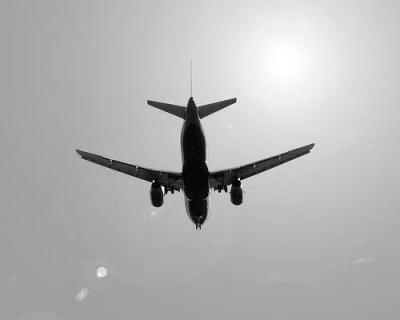}
    \includegraphics[width=\columnwidth]{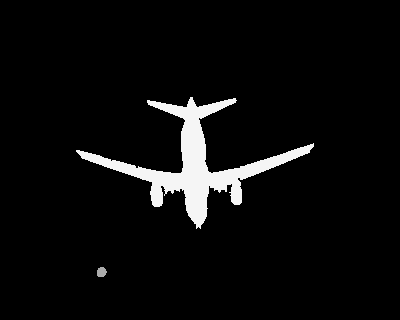} \centering D: Airplane
\end{minipage}
   \caption{Comparison with the mesh-based implementation of Renka. We compare segmentations obtained using the present particle-based Sobolev gradient approximation with the fully accurate mesh-based method of Renka~\cite{Renka:2009}. We use the complete set of all four example images considered by Renka (A--D). The raw images are shown in the top row, the final segmentation masks in the bottom row. We refer to Ref.~\cite{Renka:2009} for the results obtained with the mesh-based method. In all cases, we initialize the present algorithm using a single rectangular region. Cases (A--C) use a piecewise constant region intensity model, (D) a piecewise smooth one. Curvature regularization is used in all cases with (A) $\lambda = 0.01$, (B) $\lambda = 0.04$, (C) $\lambda = 0.04$, (D) $\lambda = 0.05$.}\label{fig:comp_renka}
\end{figure}

Next, we compare the present particle approximation with a fully accurate mesh-based Sobolev solver, both in terms of the solutions found and in terms of computational cost. In Fig.~\ref{fig:comp_renka} we consider the same four test images as were used to benchmark the mesh-based level-set implementation of Renka~\cite{Renka:2009}. The results obtained with the present particle method are visually indistinguishable from the results shown in Ref.~\cite{Renka:2009}, and even seem to be slightly better for the Airplane image. The present particle approximation does hence not seem to have a detectable adverse effect on the solution quality compared with a fully resolved mesh-based level-set method. The present implementation using particle--particle interactions, however, leads to significant computational savings, as shown in Table \ref{tab:comp_renka_itime}. Compared with the full-grid method, the present algorithm is between 3.6 and 17 times faster. Compared with the efficient narrow-band level-set implementation, the present particle method is still between 2.8 and 10 times faster. The present method is also easier to implement, as it does not require programming an additional Poisson solver.
 
 \begin{table}[htbp]
  \centering
 \begin{tabular}{|c|c|c|c|c|}
  \hline
  Image & Gradient type  & Ref.~\cite{Renka:2009} full grid & Ref.~\cite{Renka:2009} narrow band & Present   \\
  \hline
  \hline
  TwoCells & $L^{2}$ &  0.04 & 0.04 & 0.007 \\
   & Sobolev & 0.08 & 0.07  & 0.022 \\
  \hline
  TwoObjects & $L^{2}$ & 0.04 & 0.03  & 0.006 \\
   & Sobolev & 0.06 & 0.07  & 0.007 \\
  \hline
  Horseshoe & $L^{2}$ & 0.35 &  0.09  & 0.033 \\
   & Sobolev & 0.57 & 0.21  & 0.033 \\
  \hline
  Airplane & $L^{2}$ & 0.40 & 0.10  & 0.078 \\
   & Sobolev & 0.66 & 0.22  &  0.078 \\
  \hline
 \end{tabular}
 \caption{Runtime comparisons (in seconds per iteration) of the present particle-based implementation as compared with those reported by Renka for a mesh-based level-set implementation of Sobolev gradients~\cite{Renka:2009}.}\label{tab:comp_renka_itime}
 \end{table}

One case where Sobolev gradients are popular is for segmenting very noisy images, where the amount of regularization that would be required in an $L^2$ flow is so large that it would destroy the solution~\cite{Sundaramoorthi:2007,Sundaramoorthi:2008}. In Fig.~\ref{fig:Sobolev_cells} we show an example of such as case, a microscopy image of fluorescently labeled cells at low signal-to-noise ratio (SNR). The example illustrates that the present implementation retains the known property that Sobolev gradients keep the contour smooth despite the noise, resulting in less noise-sensitive segmentations, as shown in Fig.~\ref{fig:Sobolev_cells}B. The $L^2$-gradient flow requires strong curve-length penalization in order to sufficiently regularize the contour (here, \mbox{$\lambda=3$}). However, this fails in the present example, as the regions collapse when the regularization parameter is increased to that level, as shown in Fig.~\ref{fig:Sobolev_cells}A. For the same $\lambda$, this is not the case for the Sobolev flow (Fig.~\ref{fig:Sobolev_cells}B). The present example is hence impossible to segment using a piecewise constant image model with $L^2$-type gradients.
      
\begin{figure}
\centering
\begin{minipage}{0.45\textwidth} 
  \includegraphics[width=\columnwidth]{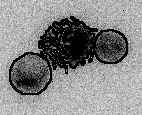} \centering A: $L^2$ gradient result
  \label{fig:sobolev_cells_with_L2}
\end{minipage}
\begin{minipage}{0.45\textwidth} 
  \includegraphics[width=\columnwidth]{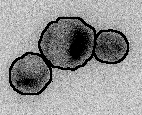} \centering B: Sobolev gradient result
  \label{fig:sobolev_cells_with_sobolev}
\end{minipage}
\caption{Sobolev gradients allow segmenting very noisy images, as shown here for fluorescently labeled cells using the piecewise constant energy from Ref.~\cite{Cardinale:2012}. Region fusions are disallowed and the regions are initialized at local intensity maxima after image blurring. (A) Final RC segmentation when using $L^2$-type gradient approximations. (B) Final RC segmentation when using Sobolev gradient approximations. In both cases \mbox{$\lambda=3$}.}\label{fig:Sobolev_cells} 
\end{figure}

 \begin{figure}[htbp]
  \centering
  \begin{minipage}{0.3\textwidth}
    \centering \includegraphics[scale=0.85]{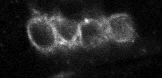} \\ \centering A: raw data
  \end{minipage}
  \begin{minipage}{0.3\textwidth}
    \includegraphics[scale=0.85]{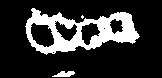}  \\ \centering B: Sobolev, $\lambda=1$, iter.~$23$
\end{minipage}
  \begin{minipage}{0.3\textwidth}
    \includegraphics[scale=0.85]{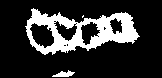}  \\ \centering C: Sobolev, $\lambda=1$, iter.~$24$
  \end{minipage}
   \vspace{1em}
   
  \begin{minipage}{0.3\textwidth}
    \includegraphics[scale=0.85]{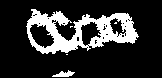}  \\ \centering D: $L^2$, $\lambda=1$
\end{minipage}
  \begin{minipage}{0.3\textwidth}
    \includegraphics[scale=0.85]{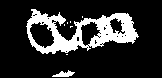}  \\ \centering E: $L^2$, $\lambda=2$
  \end{minipage}
  \begin{minipage}{0.3\textwidth}
    \includegraphics[scale=0.85]{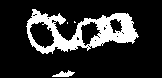}  \\ \centering F: $L^2$, $\lambda=3$
  \end{minipage}

   \vspace{1em}
   
    \begin{minipage}{0.3\textwidth}
      \includegraphics[scale=0.85]{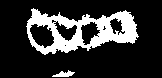}  \\ \centering G: Sobolev, $\lambda=1$
  \end{minipage}
  \begin{minipage}{0.3\textwidth}
    \includegraphics[scale=0.85]{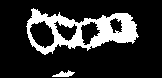}  \\ \centering H: Sobolev, $\lambda=2$
  \end{minipage}
   \begin{minipage}{0.3\textwidth}
    \includegraphics[scale=0.85]{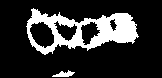}  \\ \centering I: Sobolev, $\lambda=3$
  \end{minipage}

  \caption{Sobolev gradient flows are less sensitive to the choice of regularization parameter. Comparison of segmentation results for a noisy part of a {\it{Drosophila melanogaster}} wing disc tissue using the $L^{2}$-type gradient approximation (D--F) and the Sobolev gradient approximation (G--I). The RC algorithm is used with a piecewise smooth energy and a Poisson noise model~\cite{Cardinale:2012}. The initial segmentation is obtained by Otsu thresholding. The final foreground region is the set union of all foreground regions found by RC. (A) Raw image with cell membranes fluorescently labeled. Image courtesy of Prof.~Christian Dahmann, Technical University of Dresden. (B) Result using Sobolev gradients until spatial convergence. Oscillations are detected at iteration 23 and the algorithm switches to $L^2$ gradients, leading to a less smooth result after the next iteration (C). (D) RC result with $L^{2}$ gradients and length regularization $\lambda=1$. (E) RC result with $L^{2}$ gradients and $\lambda=2$. (F) RC result with $L^{2}$ gradients and $\lambda=3$. (G) RC result with Sobolev gradients and $\lambda=1$. (H) RC result with Sobolev gradients and $\lambda=2$. (I) RC result with Sobolev gradients and $\lambda=3$.}
  \label{fig:wing_noise}
\end{figure}

Sobolev gradients precondition the optimization problem and lead to less ill-posed inverse problems~\cite{Sundaramoorthi:2007,Sundaramoorthi:2008}. 
This renders the result less sensitive to the choice of regularization constant $\lambda$, i.e., giving more weight to the image data and less weight to the Bayesian prior (see Methods section). This is desirable since it frees the user of the tedious parameter tuning of finding a ``good'' $\lambda$ for a given image. We illustrate that the present fast implementation retains the property of being robust against the choice of $\lambda$. We do so by comparing results obtained with the RC algorithm with $L^2$-type and Sobolev gradient approximations for different values of the regularization constant (higher $\lambda$ means stronger regularization and more weight to the prior). The results are shown in Fig.~\ref{fig:wing_noise} for fluorescently labeled cell membranes of a small group of four cells in a fly wing disk, acquired at low SNR (raw data courtesy of Prof.~Ch.~Dahmann, TU Dresden). The fluorescence signal varies inhomogeneously across the tissue (see Fig.~\ref{fig:wing_noise}A), which is why we use a piecewise smooth image model in this case~\cite{Cardinale:2012}. We also assume Poisson noise in the image, since shot noise is the dominant noise source in confocal fluorescence microscopy. 
Figure~\ref{fig:wing_noise}A shows the raw data. After 23 iterations, the RC algorithm with Sobolev gradient converged to a point where the contour oscillates around a local energy minimum (Fig.~\ref{fig:wing_noise}B). The algorithm hence switches to $L^2$ gradients, leading to a significantly less smooth contour after the subsequent iteration, i.e., after one step of $L^2$-gradient descent (Fig.~\ref{fig:wing_noise}C). Nevertheless, the result is still smoother than when using $L^2$ gradients from the beginning (Fig.~\ref{fig:wing_noise}D; 39 iterations required to converge), indicating that the Sobolev gradient flow converged to a different local minimum than the $L^2$ flow. When increasing the regularization constant $\lambda$, the $L^2$ results become smoother, but image features start to be missed, leading in to the formation of a ``hole'' in membrane of the second cell from the left (Figs.~\ref{fig:wing_noise}E--F). It is hence very important for $L^2$ flows that the user properly tunes $\lambda$ by trial and error in order to obtain a topologically correct segmentation. The final segmentations when using Sobolev gradients are less sensitive to changes in $\lambda$ and yield the correct topology of the segmentation mask in all tested cases (Figs.~\ref{fig:wing_noise}G--I). When using Sobolev gradients, the contour is also less sensitive to noise and good segmentation results are obtained already for smaller regularization constants than when using $L^2$-type gradients. This is beneficial in practice, as it frees to user of some of the tedious parameter tuning involved in finding a good $\lambda$ for a given image. The present particle approximation retains this known property of Sobolev flows. The per-iteration CPU time is 0.01\,s when using $L^2$ gradients and 0.02\,s for Sobolev gradients.

\begin{figure}[htbp]
  \centering
  \begin{minipage}{0.24\textwidth}
    \includegraphics[scale=0.8]{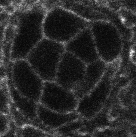}  \\ \centering A: raw data
  \end{minipage}
  \begin{minipage}{0.24\textwidth}
    \includegraphics[scale=0.8]{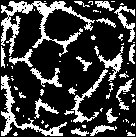} \\ \centering B: $L^2$, $\lambda=0.5$
  \end{minipage}
   \begin{minipage}{0.24\textwidth}
    \includegraphics[scale=0.8]{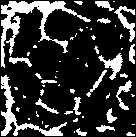}  \\ \centering C: $L^2$, $\lambda=1.5$
  \end{minipage}
  \begin{minipage}{0.24\textwidth}
    \includegraphics[scale=0.8]{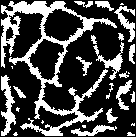}  \\ \centering D: Sobolev, $\lambda=0.5$
  \end{minipage}
  \caption{Sobolev gradients relax the tradeoff between solution smoothness and completeness. Comparison of segmentation results for filaments in a cyst in {\it{Drosophila melanogaster}} using the $L^{2}$-type gradient approximation (B--C) and the Sobolev gradient approximation (D). The RC algorithm is used with a piecewise smooth energy and a Poisson noise model~\cite{Cardinale:2012}. The initial segmentation is obtained by Otsu thresholding. (A) Raw data with spectrin fluorescently labeled. Images courtesy of Dr.~Guillaume Salbreux, Max Planck Institute for the Physics of Complex Systems, Dresden. (B) RC result with $L^{2}$ gradients and $\lambda=0.5$. (C) RC result with $L^{2}$ gradients and $\lambda=1.5$ (D) RC result with Sobolev gradients and $\lambda=0.5$.}
  \label{fig:spec}
\end{figure}

In any estimation problem, there is a fundamental tradeoff between completeness (fitting the data as completely as possible) and smoothness (generalization; not over-fitting the noise in the data). This is also true in image segmentation. The regularization parameter $\lambda$ allows the user to tune this tradeoff. It is known that the increased robustness of Sobolev gradients against the choice of $\lambda$ comes from the fact that the optimization problem is pre-conditioned in a way that relaxes this tradeoff~\cite{Sundaramoorthi:2007,Sundaramoorthi:2008}. We illustrate next that the present particle approximation shares this property of relaxing the tradeoff between contour smoothness and not missing weak image features (see Fig.~\ref{fig:spec}). The raw data in Fig.~\ref{fig:spec}A shows a part of a cyst in a fruit fly with the cytoskeletal protein spectrin fluorescently labeled. The RC segmentation results using $L^2$-type gradients are shown in Figs.~\ref{fig:spec}B--C. For a small $\lambda$, the contour is non-smooth, as it delineates noise (Fig.~\ref{fig:spec}B; over-fitting). For even smaller $\lambda$ the $L^2$ gradient flow does not converge due to insufficient regularization. This is hence the most-connected segmentation one can obtain on this image when using $L^2$ gradients. Increasing the regularization constant $\lambda$ makes the contour smoother, but causes filaments to break and to go missing, as shown in Fig.~\ref{fig:spec}C. We use a piecewise smooth image model here, which is able to handle intensity variations within a foreground region. The missing and broken filaments in the center of the image in Fig.~\ref{fig:spec}C are hence not a problem of photometry, but are caused by the $L^2$ gradient flow being trapped in a local minimum due to noise. This is the reason for the above-mentioned tradeoff between smoothness and completeness. It is known that this tradeoff is relaxed when using Sobolev gradients, leading to smoother and more connected segmentations already for low regularization constants~\cite{Sundaramoorthi:2007,Sundaramoorthi:2008}, as confirmed in Fig.~\ref{fig:spec}D. Albeit in the present case the segmentation remains unsatisfactory from a biological point of view, the present fast particle approximation seems to preserve the property of Sobolev flows to relax the tradeoff between smoothness and completeness of a segmentation. For $\lambda=0.5$, the per-iteration CPU time was 0.04\,s for $L^2$ gradients and 0.08\,s for Sobolev gradients. The ratios for other $\lambda$'s were similar.
   
 \begin{figure}[htbp]
  \centering
  \begin{minipage}{0.48\textwidth}
    \includegraphics[scale=0.28]{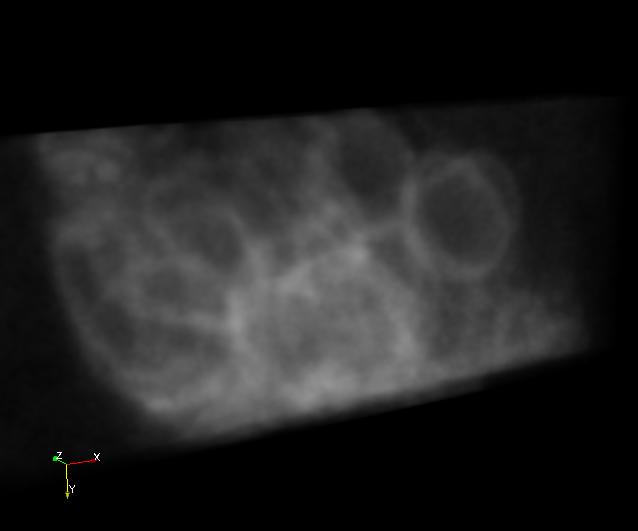}\\ \centering A: raw data, volume rendering
  \end{minipage}
  \begin{minipage}{0.48\textwidth}
    \includegraphics[scale=0.28]{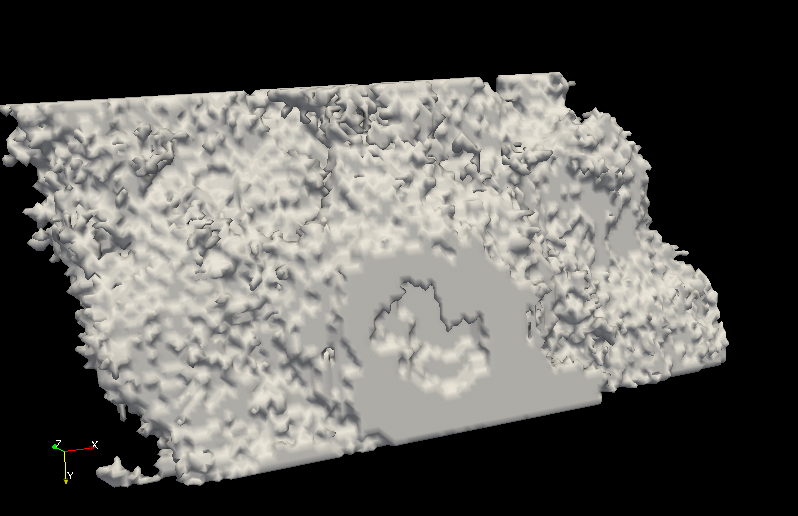}\\ \centering B: $L^2$, $\lambda=0.1$
  \end{minipage}
  \vspace{1em}
  
    \begin{minipage}{0.48\textwidth}
    \includegraphics[scale=0.28]{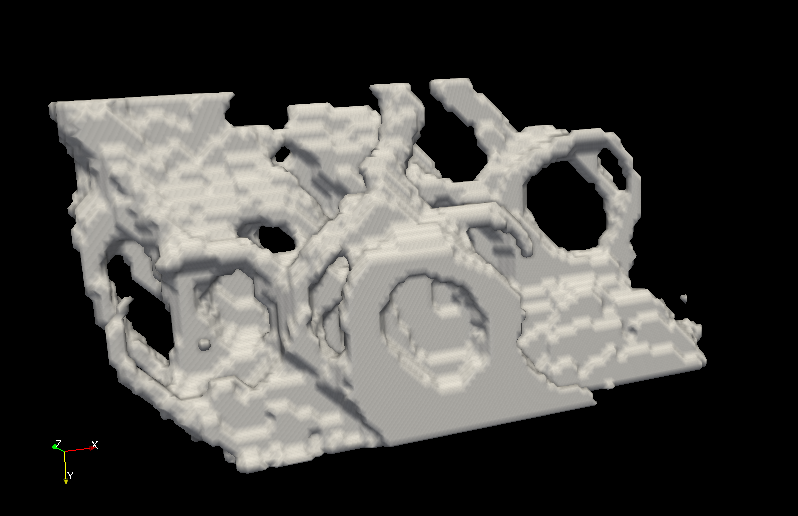}\\ \centering C: $L^2$, $\lambda=5$
  \end{minipage}
  \begin{minipage}{0.48\textwidth}
    \includegraphics[scale=0.28]{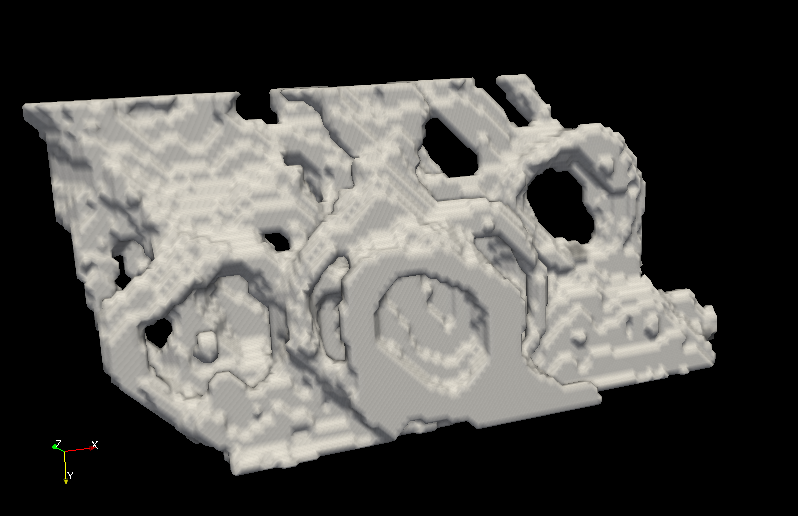} \\ \centering D: Sobolev, $\lambda=0.01$
  \end{minipage}
    \vspace{1em}
  \caption{Extension to 3D and segmentation using weaker regularization. Comparison of segmentation results for a 3D confocal microscopy stack of a noisy part of the {\it{Drosophila melanogaster}} wing disc using $L^2$-type gradients (B--C) and Sobolev gradients (D). The RC algorithm is used with a piecewise smooth energy and a Poisson noise model~\cite{Cardinale:2012}. The initial segmentation is obtained by Otsu thresholding. (A) Volume rendering of the 3D raw image with cell membranes fluorescently labeled. Image courtesy of Prof.~Christian Dahmann, Technical University of Dresden. (B) RC result with $L^{2}$ gradients and $\lambda=0.1$. (C) RC result with $L^{2}$ gradients and $\lambda=5$. (D) RC result with Sobolev gradients and $\lambda=0.01$. When using $L^2$ gradients, the algorithm did not converge for $\lambda=0.01$ due to insufficient regularization.}
    \label{fig:wing_noise3D}
\end{figure}

We next show that the present particle-based contour representation naturally extends to 3D images and allows segmentation using much less regularization than what would be required by an $L^2$ flow to generate comparable results. This is important in practice, because the regularizer is often chosen {\it{ad hoc}} with no biological meaning. One hence wishes to keep $\lambda$ as small as possible in order to limit the regularization bias in the result, but still $\lambda$ has to be chosen large enough so the solution does not over-fit the noise in the image. Figure~\ref{fig:wing_noise3D} shows a 3D example with a piecewise smooth image model. Using Sobolev gradients yields a smoother and less noisy segmentation with smaller $\lambda$, as shown in Fig.~\ref{fig:wing_noise3D}D. For the same value of $\lambda=0.01$, the $L^2$ gradient descent does not converge due to insufficient regularization. The smallest regularization for which the $L^2$ flow produces a stable result is $\lambda=0.1$. This result is shown in Fig.~\ref{fig:wing_noise3D}B. It is completely irregular and dominated by noise. In order to achieve a result that is qualitatively comparable to that in Fig.~\ref{fig:wing_noise3D}D, the $L^2$ flow requires a 500-fold(!) stronger regularization than the Sobolev flow (Fig.~\ref{fig:wing_noise3D}C). This massive regularization, however, introduces a visible bias in the final result. Some of the loops and geometric features in the tissue are broken (Fig.~\ref{fig:wing_noise3D}C). At $\lambda=0.1$, RC with Sobolev gradients requires 359 iterations to converge, with $L^2$ gradients 896 iterations. The per-iteration CPU time was 8\,s when using $L^2$ gradients and 11\,s when using Sobolev gradients. Sobolev gradients hence segment this image about twice faster than $L^2$ gradients. 
       
\begin{figure}
  \begin{minipage}{0.48\textwidth}
    \includegraphics[width=\columnwidth]{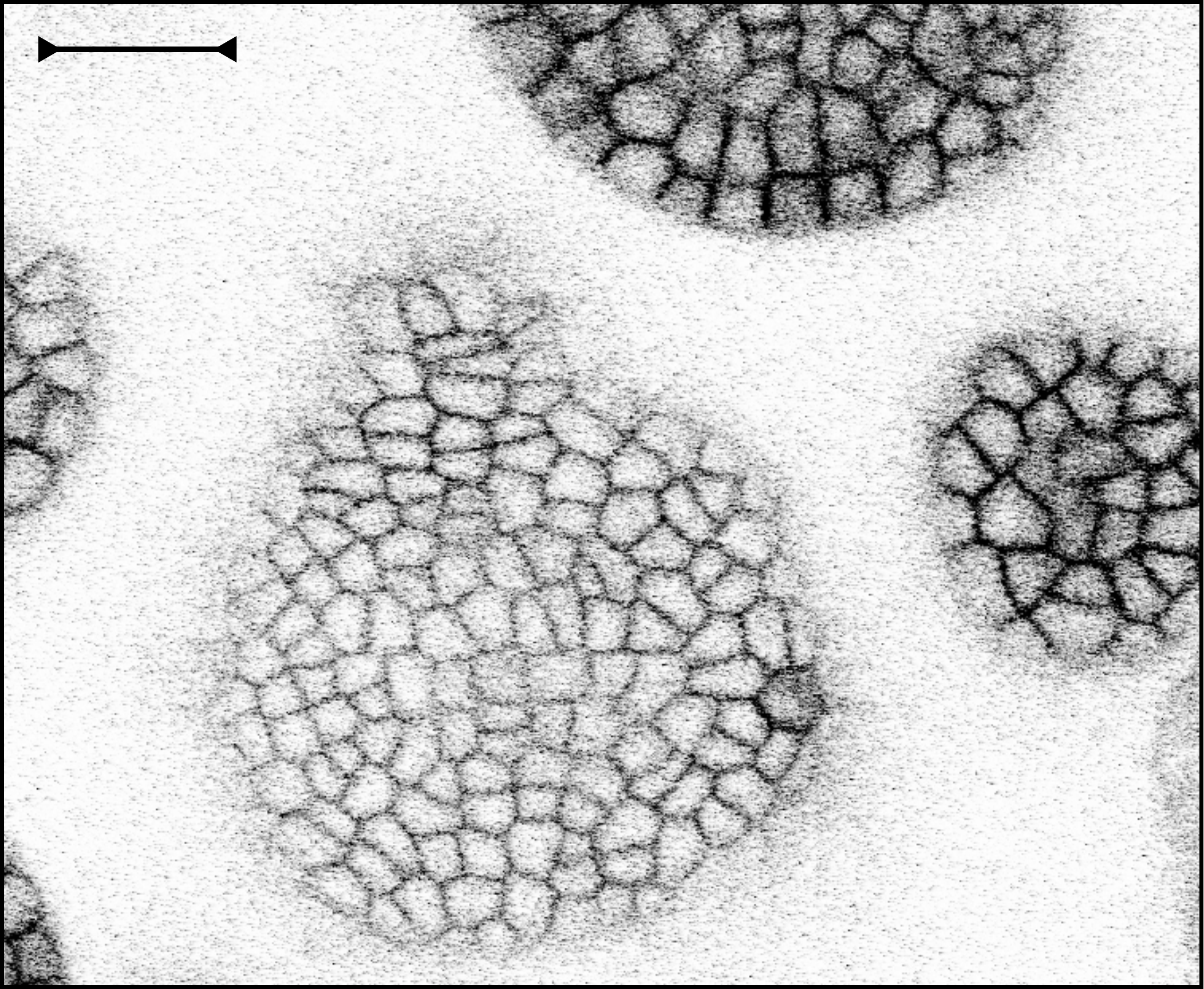}   \\ \centering A: raw data, scale bar: 20\,$\mu$m
  \end{minipage} 
  \begin{minipage}{0.48\textwidth}
    \includegraphics[width=\columnwidth]{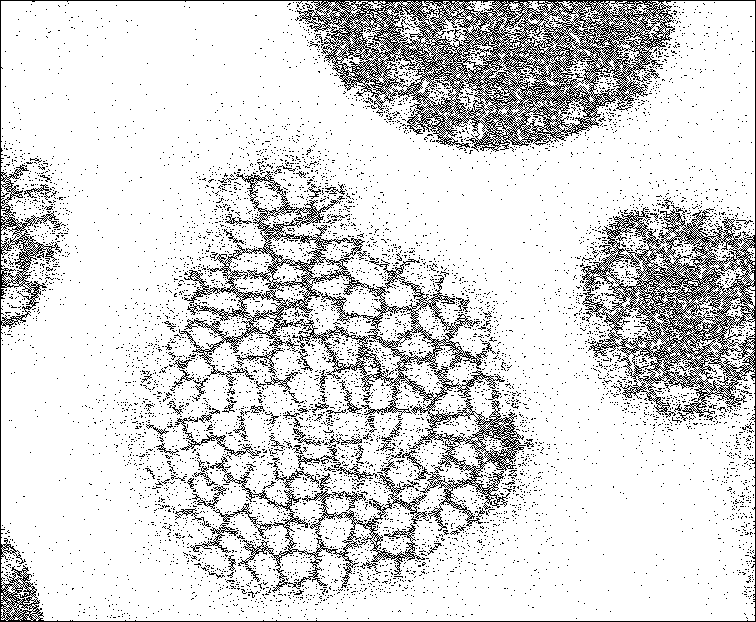}  \\ \centering B: Otsu thresholding
  \end{minipage} 
  
\vspace{5mm}
  \begin{minipage}{0.48\textwidth}
    \includegraphics[width=\columnwidth]{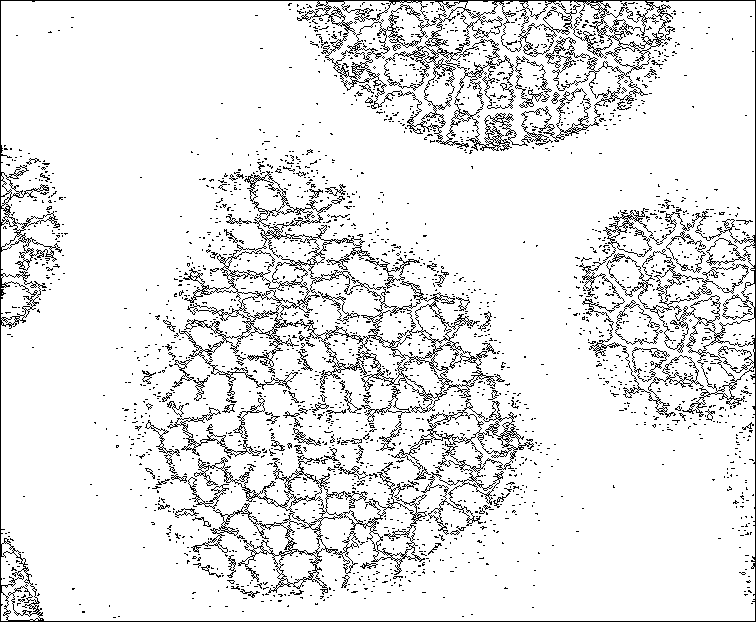}  \\ \centering C: $L^2$ gradient result, $\lambda=3$
  \end{minipage} 
  \begin{minipage}{0.48\textwidth}
    \includegraphics[width=\columnwidth]{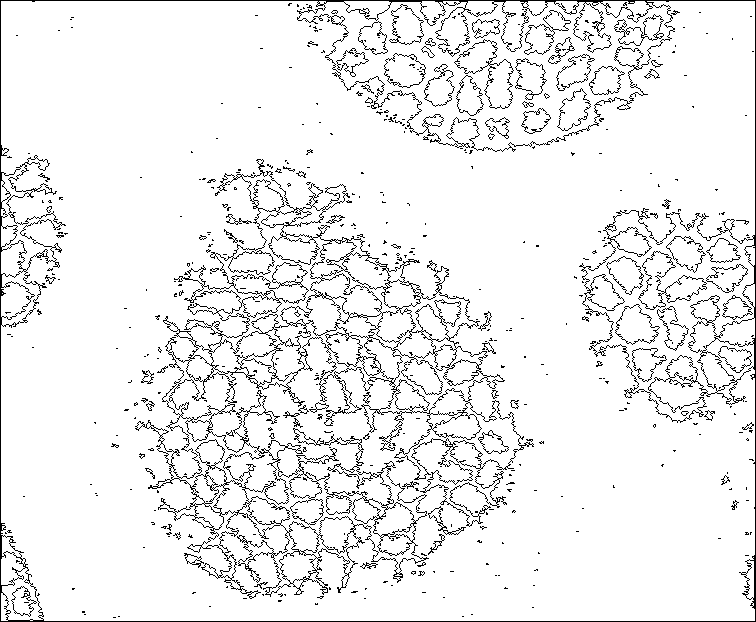}  \\ \centering D: Sobolev gradient result, $\lambda=3$
  \end{minipage} 
  \caption{Sobolev gradients are less prone to converge to noisy local minima, as shown here for shoot apical meristems of {\it{Arabidopsis thaliana}} labeled with a fluorescent protein that localizes to the plasma membrane. (A) Raw image of size $756\times 622$\,pixels (source: \cite{Liu:2010}). 
       (B) Initial segmentation using Otsu thresholding. (C) Segmentation result using $L^2$ gradients.  (D) Segmentation result using approximated Sobolev gradients. In both cases $\lambda=3$, hence keeping the energy landscape the same.}\label{fig:plant}
\end{figure}

Another known property of Sobolev gradients is that they lead to final segmentations with less ``holes'', as the flow can overcome small local minima that are induced by noise. Figure \ref{fig:plant} illustrates that the present particle approximation retains this property. The figure shows an image of fluorescently labeled plant tissue at low SNR (data: \cite{Liu:2010}). Even for the same $\lambda=3$, the Sobolev flow leads to contours that are more connected and less broken (Fig.~\ref{fig:plant}D) than those obtained by the $L^2$ flow (Fig.~\ref{fig:plant}C). Since the same $\lambda$ is used in both cases, this can only be because the Sobolev flow converges to a different, less noisy local minimum. This present implementation hence retains this property. RC with $L^2$ and gradients needed 106 iterations (254\,s) to converge, with Sobolev gradients 200 iterations (549\,s). This is hence an example of where the Sobolev flow converges slower, but finds a different, better local minimum.
      
\begin{figure}
  \begin{minipage}{0.32\textwidth}
    \includegraphics[width=\textwidth,clip,trim=120px 100px 100px 100px]{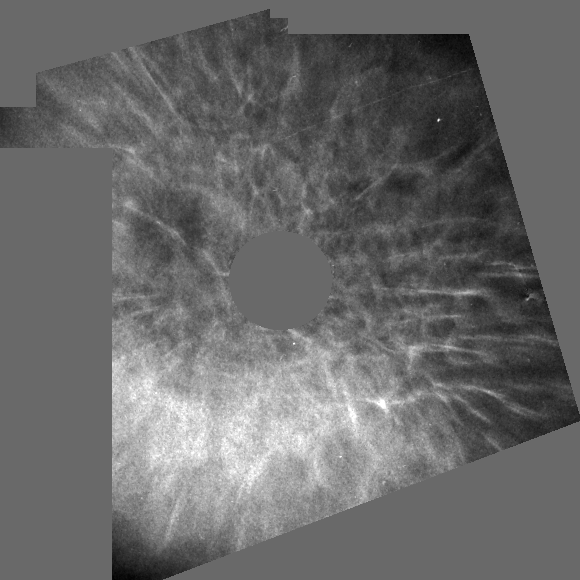}    \\ \centering A
  \end{minipage}
  \begin{minipage}{0.32\textwidth}
    \includegraphics[width=\textwidth,clip,trim=120px 100px 100px 100px]{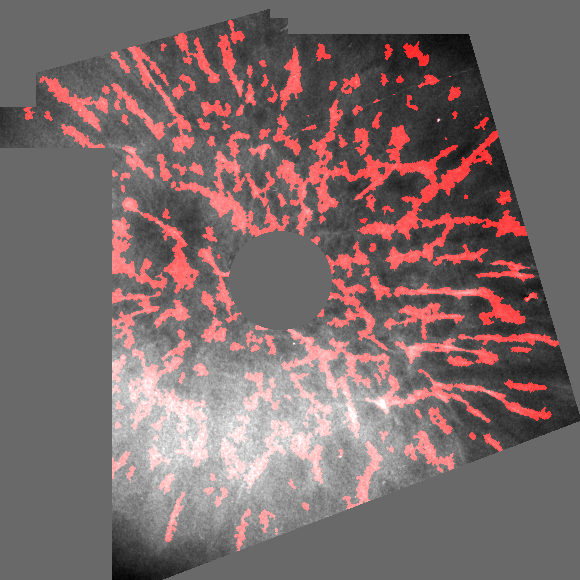}      \\ \centering D
  \end{minipage}
  \begin{minipage}{0.32\textwidth}
    \includegraphics[width=\textwidth,clip,trim=120px 100px 100px 100px]{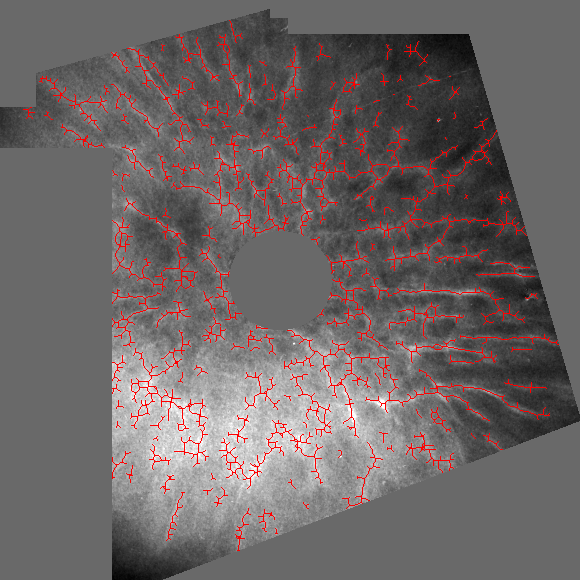}      \\ \centering G
  \end{minipage}

\vspace{5mm}
  \begin{minipage}{0.32\textwidth}
    \includegraphics[width=\textwidth,clip,trim=120px 100px 100px 100px]{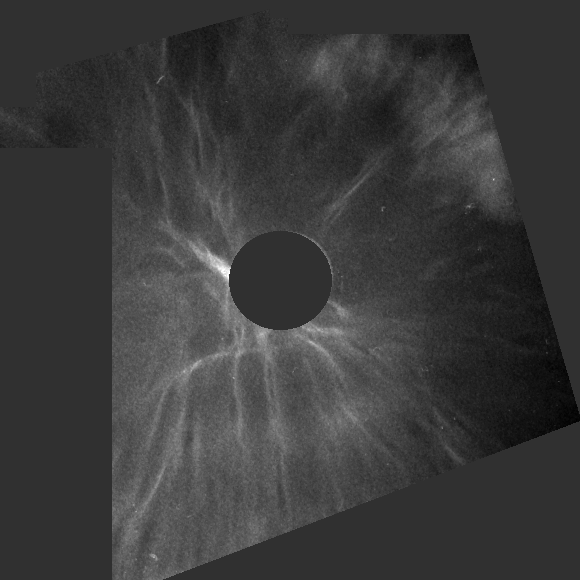}      \\ \centering B
  \end{minipage}
  \begin{minipage}{0.32\textwidth}
    \includegraphics[width=\textwidth,clip,trim=120px 100px 100px 100px]{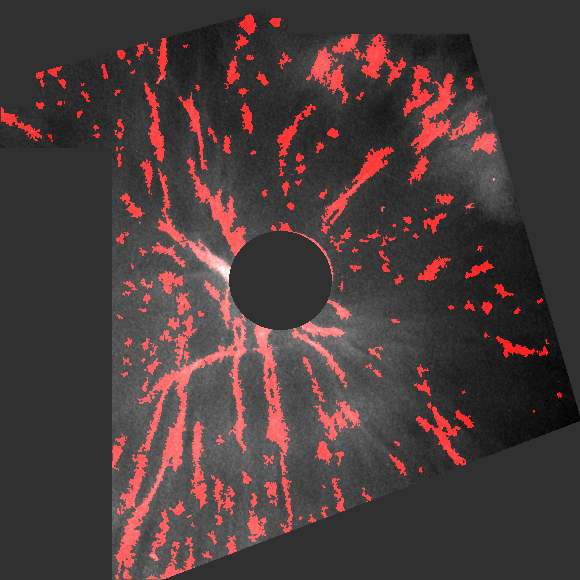}      \\ \centering E
  \end{minipage}
  \begin{minipage}{0.32\textwidth}
    \includegraphics[width=\textwidth,clip,trim=120px 100px 100px 100px]{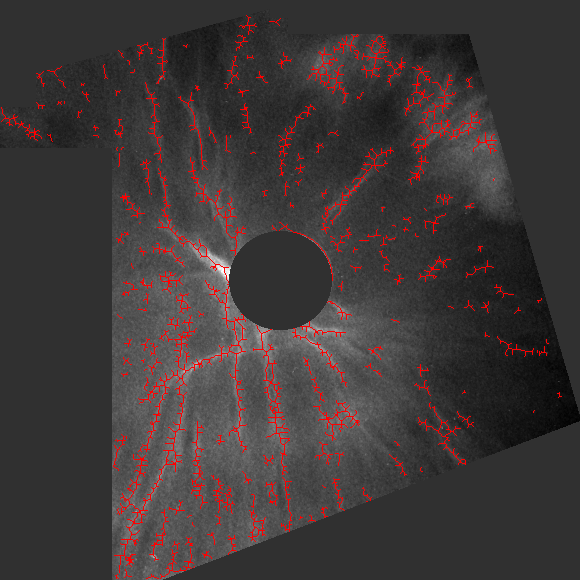}      \\ \centering H
  \end{minipage}

\vspace{5mm}
  \begin{minipage}{0.32\textwidth}
    \includegraphics[width=\textwidth,clip,trim=120px 100px 100px 100px]{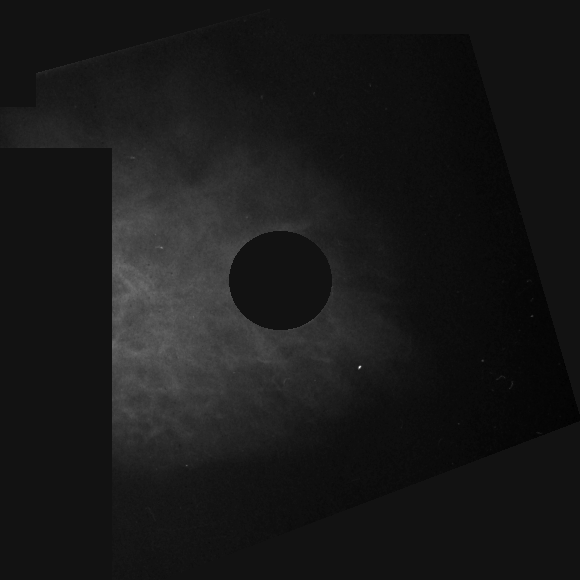}      \\ \centering C
  \end{minipage}
  \begin{minipage}{0.32\textwidth}
    \includegraphics[width=\textwidth,clip,trim=120px 100px 100px 100px]{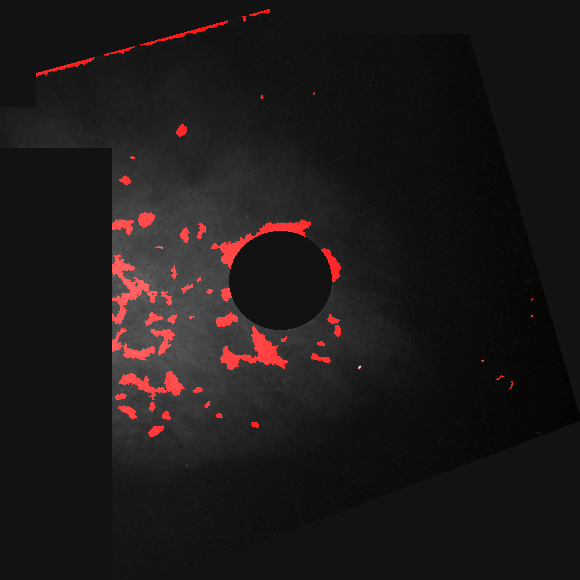}      \\ \centering F
  \end{minipage}
  \begin{minipage}{0.32\textwidth}
    \includegraphics[width=\textwidth,clip,trim=120px 100px 100px 100px]{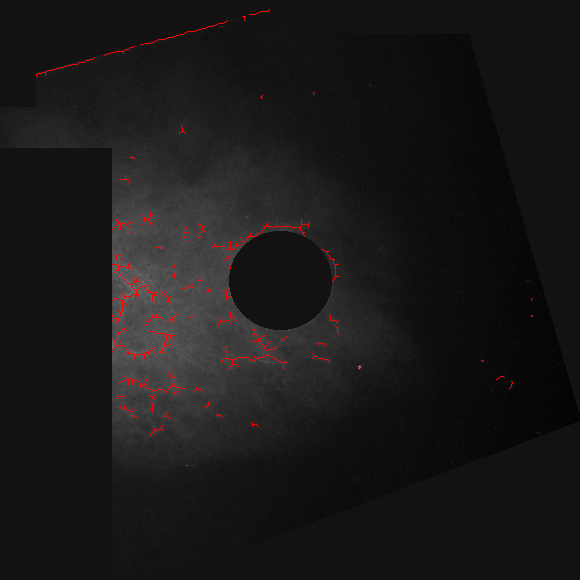}      \\ \centering I
  \end{minipage}
\caption{Sobolev gradients require no additional regularization on images with little noise, as shown here for segmentations of proton emission patterns without additional regularization. (A--C) Raw data (images: Josefine Metzkes, Bussmann group, Helmholtz Center Dresden Rossendorf). (D--F) RC segmentations using a piecewise smooth image model~\cite{Cardinale:2012} and Sobolev gradients with \mbox{$\lambda=0$}. (G--I) Skeletonization of the segmentation results in order to extract filament patterns.}\label{fig:laser}
\end{figure} 

Since Sobolev flows require less additional regularization, they are also popular to process images that contain very little or no noise. In those cases, Sobolev flows sometimes allows segmenting the image with no additional regularization at all. We present such an example in Fig.~\ref{fig:laser}. It is not a microscopy example, because microscopy images always contain significant noise. Instead, the image shows proton emission patterns that occur when a high-energy laser beam hits a block of metal (data: Bussmann group, Helmholtz Center Dresden Rossendorf). We initialize the RC algorithm with small bubbles around local intensity maxima. Since intensities are varying within filament patterns, we use a piecewise smooth energy model~\cite{Cardinale:2012}. The images are almost noise-free, and the present fast particle-Sobolev flow allows segmenting them with $\lambda=0$, i.e., without any additional regularization. This absence of regularization maximizes the detection ability for dark filaments, a property that is retained by the present particle approximation. However, the final result is sensitive to the initialization, and the topological prior that a region is a connected component is necessary in this case~\cite{Cardinale:2012}.

In summary, the examples shown here confirm that the present approximation of Sobolev gradients using local particle interactions qualitatively retains the known qualitative properties of Sobolev flows and does not seem to have a detectable adverse effect on solution quality. However, the local character of the computations and the simple implementation as particle--particle interactions provide savings in both runtime and programmer effort. The use of Sobolev gradients preconditions the energy-minimization problem. This leads to smoother results that are less sensitive to noise and to the value of the regularization constant $\lambda$. The Sobolev gradient flow frequently converges to a different local minimum than the $L^2$-type gradient flow and requires less regularization.  

\section*{Methods}\label{sec:methods}
\subsection*{Bayesian image models}
A Bayesian model is determined by a likelihood $p(I|\Gamma,\vec{\theta})$ and a prior $p(\Gamma)$, where in image segmentation $I$ is the image data and $\Gamma$ the segmentation contour. The vector $\vec{\theta}$ contains photometric parameters such as region intensities. The posterior probability density function is obtained using Bayes' formula: 
\begin{equation}
  \label{eq:Bayes_formula}
  p(\Gamma,\vec{\theta}|I)=\frac{p(I|\Gamma,\vec{\theta})\cdot p(\Gamma,\vec{\theta})}{p(I)}.
\end{equation}

The likelihood expresses how likely it is to observe the measured image $I$ given a certain segmentation $\Gamma$ and parameters $\vec{\theta}$. The likelihood therefore formalizes the image-formation model. The image-formation model describes the mapping between a ground-truth object state $(\Gamma_{\text{GT}}, \vec{\theta}_\text{GT})$ and an expected image $J$ conditional on that state. For many image-formation models, $\vec{\theta}$ is jointly determined by the model and by $I$. For example, in many popular models the estimated intensity $\theta_i$ of a region $i$ is equal to the mean intensity of the area enclosed by that region's contour $\Gamma_i$ in $I$. We therefore mostly omit $\vec{\theta}$ in our notation.

Generating $J$ from $\Gamma$ is called the {\em{forward problem}}. It amounts to simulating the expected (i.e., noise-free) image under a given segmentation. In fluorescence microscopy, the forward model is linear: 
\begin{equation}
\label{eq:image_formation}
J = s_{\vec{\theta}}(\Gamma) * K 
\end{equation}
and amounts to a convolution of the expected intensity distribution $s_{\vec{\theta}}(\Gamma)$ with the point-spread function (impulse-response function) $K$ of the microscope. The function $s_{\vec{\theta}}$ is called the image-generating function. It assigns an expected intensity to each pixel according to the image-formation model. Frequently used image-generating functions are the piecewise constant approximation, assigning a constant intensity $\theta _i$ to all pixels enclosed by contour $\Gamma _i$, and the piecewise smooth approximation, assigning shaded intensities within regions. Estimating the region intensities $\vec{\theta}$ (and their distribution inside regions) from the image data is called the {\em{photometric estimation problem}}. Finding the contours $\Gamma_i$ constitutes the {\em geometric estimation problem}. Estimating $\Gamma$ and $\vec{\theta}$ from given $I$, $K$, and $s_{\vec{\theta}}$ is an {\em{inverse problem}} that is addressed using Bayesian inference.

Prior terms measure how likely a certain segmentation $\Gamma$ is, independent of the observed image. The most popular prior term penalizes the  contour length $|\Gamma|$, favoring short (and hence smooth) region boundaries~\cite{Mumford:1989}. 
Other priors may include global shape characteristics and penalize deviations of the segmented shape from a template shape~\cite{Leventon:2000,Raviv:2004}. 

Both the likelihood (i.e., the forward model) and the Bayesian prior can be used to include prior knowledge into the image-analysis algorithm. We hence define {\em prior knowledge} more general than just the {\em Bayesian prior}.

\subsection*{Posterior maximization}
In Bayesian (i.e., model-based) image segmentation, the estimation problem of finding $\Gamma$ and $\vec{\theta}$ given $I$, $K$, and $s_{\vec{\theta}}$ is  formulated as a maximum-a-posteriori (MAP) problem
\begin{equation}
  \max_{\Gamma,\vec{\theta}}{ p(\Gamma, \vec{\theta}|I)} \, . 
\end{equation}
The problem is often restated as minimizing the anti-logarithm of the posterior, called the {\em{energy function}} $\mathcal{E}=-\log  p(\Gamma,\vec{\theta}| I)$. The use of a logarithm is motivated by the Boltzmann distribution and changes the product of the likelihood and the prior to a sum of energy terms. The energies corresponding to the likelihood and the prior are called the {\em{external}} and {\em{internal}} energy, respectively. The resulting energy-minimization problem then reads:
\begin{equation}
\min_{\Gamma,\vec{\theta}}\left[ -\log p(I|\Gamma,\vec{\theta}) - \log p(\Gamma,\vec{\theta}) \right] = \min_{\Gamma,\vec{\theta}}\left[ \mathcal{E}_{\textrm{external}} + \lambda \mathcal{E}_{\textrm{internal}} \right].
\end{equation}
The denominator in Bayes' formula~(\ref{eq:Bayes_formula}) drops out as a constant shift. The scalar parameter $\lambda$ is included in order to weight the prior with respect to the likelihood. The larger $\lambda$, the more weight is given to the prior and the less weight to the image data. This increases the regularization. Choosing the optimal $\lambda$ is an open problem. In practice, however, one usually wants to choose $\lambda$ as small as possible to still get a robust segmentation and give as much weight to the image data $I$ as possible. 

A plethora of minimization algorithms for the total energy $\mathcal{E}=\mathcal{E}_{\textrm{external}} + \lambda \mathcal{E}_{\textrm{internal}}$ has been presented in the literature. The method of choice depends on the characteristics of $\mathcal{E}$. Gradient-based local optimization of $\mathcal{E}$ is a popular approach for non-convex $\mathcal{E}$. In a discrete space, the energy gradient becomes an energy difference $\Delta\mathcal{E}$. In order to evolve the contour along the energy gradient flow, we (only) need to be able to evaluate energy differences, corresponding to posterior ratios 
$
\frac{p(\Gamma'|I)}{p(\Gamma|I)}=\exp(-\Delta\mathcal{E})
$
for the original contour $\Gamma$ and the perturbed contour $\Gamma'$.
The quantity of interest hence is the energy difference $\Delta\mathcal{E}$ when deforming $\Gamma$ to $\Gamma'$. We assume that image noise is realized independently for each pixel. Using Bayes' formula (\ref{eq:Bayes_formula}) we then decompose $\Delta\mathcal{E}$ as
\begin{equation} 
\label{eq:deltaE}
\begin{split}
 \Delta \mathcal{E} = -\log\left(\frac{p(\Gamma'|I)}{p(\Gamma|I)}\right) = -\sum_{i=0}^{M-1} &\left(\log\prod_{x\in\Omega_i'}{p\left(I(x)|J'(x)\right)\cdot p(\Gamma')} 
  -\log\prod_{x\in\Omega_i}{p\left(I(x)|J(x)\right)\cdot p(\Gamma)}\right),
\end{split}
\end{equation}
where the images $J$ and $J'$ are computed using Eq.~(\ref{eq:image_formation}). $\Omega_i$ and $\Omega'_i$ are the regions (sets of pixels) enclosed by $\Gamma_i$ and $\Gamma_i'$, respectively. $M$ is the total number of regions. Foreground regions are defined as closed sets, whereas the background region with index 0 is an open set (see Fig.~\ref{fig:fg_simple}).

\section*{Conclusions}\label{sec:conclusions}
We have shown how Sobolev gradients can be used to drive the evolution of a discrete particle-based deformable model. The particle nature of the present method allows for a simple and efficient approximation of Sobolev gradients and dispenses with the need to implement additional global solvers. This does not only reduce the runtime of Sobolev codes, but also reduces the programmer burden during software development. Sobolev gradients are known to precondition the optimization problem in Bayesian image segmentation and hence have a number of favorable properties: they are less sensitive to noise, they require less regularization, they favor the evolution of smooth contours, and they distinguish between local contour deformations and global contour motion~\cite{Sundaramoorthi:2007,Sundaramoorthi:2008}. Sobolev gradients can be approximated from $L^2$-type gradients by convolution with a local kernel~\cite{Sundaramoorthi:2007}. In particle methods, discrete convolution amounts to particle--particle interactions~\cite{Hockney:1988,Eldredge:2002,Koumoutsakos:2005,Schrader:2010}, which can efficiently be computed if the kernel is local. Particle methods in image processing are hence naturally suited for more general gradient definitions, such as the Sobolev gradients considered here. 

The Region Competition (RC) algorithm is a local black-box (i.e., zeroth-order) optimizer for particle-based deformable models~\cite{Cardinale:2012}. Since it only requires point-wise evaluations of energy differences, which are used as discrete gradient approximations, the gradient definition can easily be changed without affecting the rest of the algorithm. In fact, Sobolev gradients can easily be plugged into any discrete optimization algorithm or convexification scheme by simply changing the way the gradient is approximated. This renders them an appealing extension to existing methods.

We have benchmarked the RC algorithm with and without Sobolev gradients on a number of artificial and real-world images, showing that the known qualitative properties of Sobolev flows are preserved by the present approximation algorithm. We have also compared the results obtained with the present approximation algorithm with results from the literature that were computed using a fully accurate mesh-based solver. The results obtained with the present particle-based scheme do not visibly differ from those obtained with mesh-based level-sets, but the present particle scheme is between 2.8 and 17 times faster than the level-set solver. Also in the present implementation, Sobolev gradients remain computationally more costly to evaluate than $L^2$-type gradients. However, they precondition the problem such that less iterations are required for the optimizer to converge to the same solution, or such that better solutions (different local minima, possibly requiring more iterations to be reached) are found. This may amortize the increased cost per iteration. 

Currently, our implementation is limited by mainly two approximations: First, we have approximated the intrinsic geodesic distance along the contour by the Euclidean distance. This limits the accuracy of the present gradient approximation if the contour is significantly curved on small length scales (\mbox{$<E$}). Future work will consider using the true geodesic distance, exploiting concepts from differential geometry as applied to particle methods on curved Riemannian manifolds~\cite{Sbalzarini:2006a,Bergdorf:2010}. Second, we have fixed the length scale $E$ of the Sobolev inner product to a user-defined value. This was required in order to keep particle--particle interactions local and hence computationally efficient. Global all-against-all interactions would incur a nominal computational complexity in $O(N^2)$ for $N$ contour particles, which is not practically feasible. In order to allow global (rigid-body) movements of the contour, it would, however, be desirable to consider all-against-all interactions in future work. The resulting $N$-body problem could be efficiently (in $O(N)$) approximated using fast multipole solvers~\cite{Greengard:1987,Greengard:1988}.

The present implementation is available from \url{http://mosaic.mpi-cbg.de} as open-source as part of the RC filter in the ITK image-processing toolkit~\cite{Ibanez:2005}, implemented in C++.

\section*{Acknowledgements}
  \ifthenelse{\boolean{publ}}{\small}{}
  We thank all members of the MOSAIC Group for the many fruitful discussions. Particular thanks go to Dr.~Gr\'{e}gory Paul for sharing his expertise in image segmentation and pointing us to the Sobolev gradient literature. We thank Prof.~Christian Dahmann (Technical University of Dresden),  Dr.~Guillaume Salbreux (Max Planck Institute for the Physics of Complex Systems, Dresden), and Dr.~Michael Bussmann (Helmholtz Center Dresden Rossendorf) for providing test images. This work was supported by the Swiss SystemsX.ch Initiative under Grant WingX and the German Federal Ministry of Research and Education (BMBF) under funding code 031A099.
 

\end{document}

%% file: closed_FG_sets.pdf_tex
\begingroup%
  \makeatletter%
  \providecommand\color[2][]{%
    \errmessage{(Inkscape) Color is used for the text in Inkscape, but the package 'color.sty' is not loaded}%
    \renewcommand\color[2][]{}%
  }%
  \providecommand\transparent[1]{%
    \errmessage{(Inkscape) Transparency is used (non-zero) for the text in Inkscape, but the package 'transparent.sty' is not loaded}%
    \renewcommand\transparent[1]{}%
  }%
  \providecommand\rotatebox[2]{#2}%
  \ifx\svgwidth\undefined%
    \setlength{\unitlength}{184bp}%
    \ifx\svgscale\undefined%
      \relax%
    \else%
      \setlength{\unitlength}{\unitlength * \real{\svgscale}}%
    \fi%
  \else%
    \setlength{\unitlength}{\svgwidth}%
  \fi%
  \global\let\svgwidth\undefined%
  \global\let\svgscale\undefined%
  \makeatother%
  \begin{picture}(1,1)%
    \put(0,0){\includegraphics[width=\unitlength]{closed_FG_sets.pdf}}%
  \end{picture}%
\endgroup%

%% file: sobolev_kernel.pdf_tex
\begingroup%
  \makeatletter%
  \providecommand\color[2][]{%
    \errmessage{(Inkscape) Color is used for the text in Inkscape, but the package 'color.sty' is not loaded}%
    \renewcommand\color[2][]{}%
  }%
  \providecommand\transparent[1]{%
    \errmessage{(Inkscape) Transparency is used (non-zero) for the text in Inkscape, but the package 'transparent.sty' is not loaded}%
    \renewcommand\transparent[1]{}%
  }%
  \providecommand\rotatebox[2]{#2}%
  \ifx\svgwidth\undefined%
    \setlength{\unitlength}{329.81881989bp}%
    \ifx\svgscale\undefined%
      \relax%
    \else%
      \setlength{\unitlength}{\unitlength * \real{\svgscale}}%
    \fi%
  \else%
    \setlength{\unitlength}{\svgwidth}%
  \fi%
  \global\let\svgwidth\undefined%
  \global\let\svgscale\undefined%
  \makeatother%
  \begin{picture}(1,0.95252989)%
    \put(0,0){\includegraphics[width=\unitlength]{sobolev_kernel.pdf}}%
    \put(0.95915941,0.05535003){\makebox(0,0)[lb]{\smash{0.5}}}%
    \put(0.09014837,0.10027409){\makebox(0,0)[lb]{\smash{0}}}%
    \put(0.06488115,0.23873436){\makebox(0,0)[lb]{\smash{0.5}}}%
    \put(0.09014837,0.3771928){\makebox(0,0)[lb]{\smash{1}}}%
    \put(0.06488115,0.51565366){\makebox(0,0)[lb]{\smash{1.5}}}%
    \put(0.09014837,0.65411455){\makebox(0,0)[lb]{\smash{2}}}%
    \put(0.06488115,0.79257239){\makebox(0,0)[lb]{\smash{2.5}}}%
    \put(0.09014837,0.93103323){\makebox(0,0)[lb]{\smash{3}}}%
    \put(0.55666565,0.05535003){\makebox(0,0)[lb]{\smash{0}}}%
    \put(0.14431805,0.05535003){\makebox(0,0)[lb]{\smash{-0.5}}}%
    \put(0.01842868,0.52677751){\color[rgb]{0,0,0}\rotatebox{90}{\makebox(0,0)[b]{\smash{$\tilde{K}E$}}}}%
    \put(0.5866525,0.00356493){\color[rgb]{0,0,0}\makebox(0,0)[rb]{\smash{$r/E$}}}%
  \end{picture}%
\endgroup%